\documentclass[letterpaper]{article} 
\usepackage{aaai23}  
\usepackage{times}  
\usepackage{helvet}  
\usepackage{courier}  
\usepackage[hyphens]{url}  
\usepackage{graphicx} 
\usepackage{amsmath, amssymb}
\usepackage{booktabs}
\usepackage[bb=dsserif]{mathalpha}
\usepackage{bm}
\usepackage{natbib}  
\usepackage{caption} 
\title{Mean-Shifted Contrastive Loss for Anomaly Detection}
\author{
    Tal Reiss,
    Yedid Hoshen
}
\affiliations{

    School of Computer Science and Engineering \\
    The Hebrew University of Jerusalem, Israel \\
    https://github.com/talreiss/Mean-Shifted-Anomaly-Detection\\
}

\begin{document}

\maketitle

\begin{abstract}
Deep anomaly detection methods learn representations that separate between normal and anomalous images. Although self-supervised representation learning is commonly used, small dataset sizes limit its effectiveness. It was previously shown that utilizing external, generic datasets (e.g. ImageNet classification) can significantly improve anomaly detection performance. One approach is outlier exposure, which fails when the external datasets do not resemble the anomalies. We take the approach of transferring representations pre-trained on external datasets for anomaly detection. Anomaly detection performance can be significantly improved by fine-tuning the pre-trained representations on the normal training images. In this paper, we first demonstrate and analyze that contrastive learning, the most popular self-supervised learning paradigm cannot be naively applied to pre-trained features.  The reason is that pre-trained feature initialization causes poor conditioning for standard contrastive objectives, resulting in bad optimization dynamics. Based on our analysis, we provide a modified contrastive objective, the Mean-Shifted Contrastive Loss. Our method is highly effective and achieves a new state-of-the-art anomaly detection performance including $98.6\%$ ROC-AUC on the CIFAR-10 dataset.
\end{abstract}

\section{Introduction}

Anomaly detection is a fundamental task for intelligent agents and has broad applications for scientific and industrial tasks. Due to the significance of the task, many efforts have been focused on automatic anomaly detection, particularly on statistical and machine learning methods. A common paradigm used by many anomaly detection methods is estimating the probability density function (PDF) of the data, given a training set of normal samples. New samples are then predicted to be normal if they are likely under the estimator, while low-likelihood samples are predicted to be anomalous. The quality of the density estimators is closely related to the quality of features used to represent the data. Classical methods used $k$-means, $k$-nearest-neighbors ($k$NN), or Gaussian mixture models (GMMs) on raw features, however, this often results in poor estimators on high-dimensional data such as images.

One way to improve PDF estimators for high-dimensional data is by representing them using effective features. Many recent methods use self-supervised features to detect anomalies. Unfortunately, anomaly detection datasets are typically small and do not include labeled anomalous samples, resulting in weak features. To overcome this issue, many previous methods suggested incorporating the use of generic, external datasets (e.g. ImageNet). As such supervision is often found off-the-shelf and does not require any further annotation, this does not present extra costs. External datasets have been used in two major ways. The first is outlier exposure (OE), which amounts to using an external dataset to simulate anomalies. Supervised techniques can then be used for anomaly detection. Although very effective, OE can completely fail when the anomalies are more similar to the normal data than to the OE dataset. 

The alternative approach, that we will follow here, advocates for pre-training representations on generic, external datasets and transferring them to the anomaly detection task. \cite{reiss2021panda} found that even a simple $k$NN anomaly detection classifier based on ImageNet pre-trained representation already outperforms nearly all self-supervised methods. Furthermore, fine-tuning the pre-trained features on the normal training data can result in significant performance improvements. Although it may appear natural that this can simply be done by initializing standard anomaly detection techniques with the pre-trained features, it is quite challenging. PANDA \cite{reiss2021panda} combined the DeepSVDD objective \cite{ruff2018deep} with pre-trained features. However, as contrastive learning approaches typically perform better than the SVDD objective, we hypothesize that combining pre-trained features with contrastive methods would achieve the best of both worlds.

We begin with the surprising result that standard contrastive methods, initialized with pre-trained weights, do not improve anomaly detection accuracy at all. An analysis of the learning dynamics reveals that this occurs as the standard contrastive loss is poorly suited for data that are concentrated in a compact subspace (which the normal data under strong pre-trained features are). We propose an alternative objective, the Mean-Shifted Contrastive (MSC) loss. The MSC loss is found to achieve better One-Class Classification (OCC) performance than the center loss (used in DeepSVDD and PANDA) and sets a new anomaly detection state-of-the-art. \textbf{Our contributions}:
\begin{enumerate}
\item We analyze the standard contrastive loss for fine-tuning pre-trained representations for OCC and show that it is poorly initialized and achieves poor performance.
\item Proposing an alternative objective, the \textit{Mean-Shifted Contrastive Loss}, and analyzing its importance for adapting features for anomaly detection. 
\item Extensive experiments demonstrating the state-of-the-art anomaly detection performance of our method (e.g. $98.6\%$ ROC-AUC on CIFAR-10).
\end{enumerate}

\section{Related Work}

\textbf{Classical anomaly detection methods:} Traditional AD methods follow three main paradigms: (i) Reconstruction e.g. principal component analysis (PCA) \cite{jolliffe2011principal} and K nearest neighbors ($k$NN) \cite{eskin2002geometric}. (ii) Density estimation e.g. Gaussian mixture models (EGMM) \cite{glodek2013ensemble}, and kernel density estimation \cite{latecki2007outlier}. (iii) One-class classification e.g. one-class support vector machine (OC-SVM) \cite{scholkopf2000support} and support vector data description (SVDD) \cite{tax2004support}.

\textbf{Self-supervised deep learning methods:} Much research was performed on learning from unlabeled data. Methods typically operate by devising an auxiliary task with automatically generated artificial labels for each image and then training a deep neural network using standard supervised techniques. Tasks include: video frame prediction \cite{mathieu2015deep}, image colorization \cite{zhang2016colorful,larsson2016learning}, puzzle solving \cite{noroozi2016unsupervised},  rotation prediction \cite{gidaris2018unsupervised}. The latter was adapted by \cite{golan2018deep, hendrycks2019using, bergman2020classification} for anomaly detection in image and tabular data. Most relevant to this work is contrastive learning \cite{chen2020simple}, which learns representations by distinguishing similar views of the same sample from other data samples. Variants of contrastive learning were also introduced to OCC. CSI \cite{tack2020csi} treats augmented input as positive samples and the distributionally-shifted input as negative samples. DROC \cite{sohn2020learning} shares a similar technical formulation as CSI without ensembling or test-time augmentation.

\textbf{Feature adaptation for one-class classification:} Hand-crafted or externally-learned representations are often suboptimal for AD. Instead, OCC methods can be initialized using pre-trained features but then adapted on OCC objectives to improve their AD accuracy. DeepSVDD \cite{ruff2018deep} pre-trains a representation encoder by autoencoding on the normal data. Several works use ImageNet pre-trained features e.g. \cite{hendrycks2019using,perera2019learning,reiss2021panda}, achieving much better results. The features are then adapted to OCC - often by the SVDD objective (e.g. DeepSVDD \cite{ruff2018deep} and PANDA \cite{reiss2021panda}). Adaptation often encounters catastrophic collapse. DeepSVDD tackles this by incorporating architectural constraints. PANDA proposed a simple early stopping approach or a more principled continual learning approach based on EWC \cite{kirkpatrick2017overcoming}.

\begin{figure*}[t]
\centering
  \centering
    \begin{tabular}{c}
\includegraphics[scale=1.3]{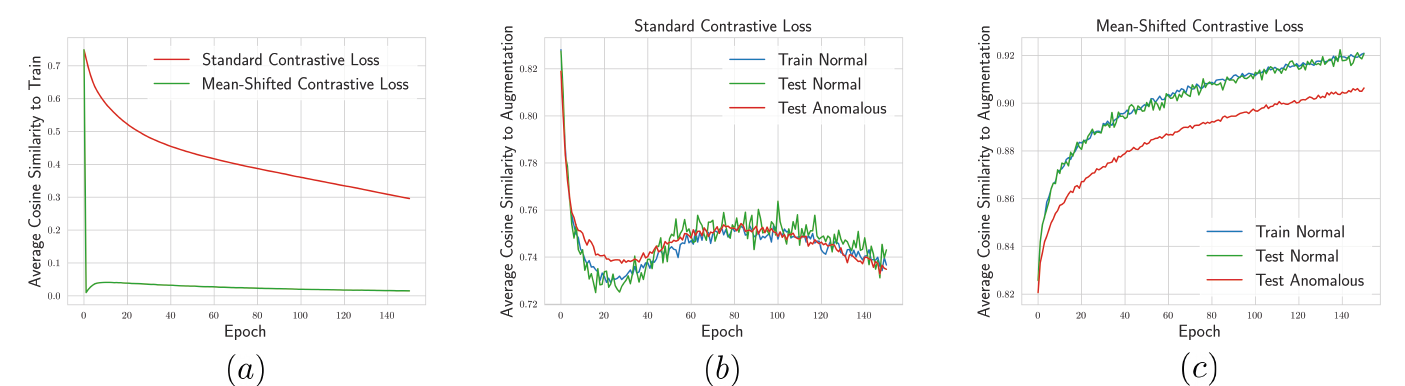}
    \end{tabular}
    \caption{CIFAR-10 "Airplane" class. Average cosine similarity between features on training set vs. training epoch. \textit{(a)} Similarity between pairs of images. Similarity between images and their augmentation for \textit{(b)} Contrastive objective  \textit{(c)} MSC objective.}
    \label{fig:fail_contrastive}

\end{figure*}

\section{Background: Learning Representations for One-Class Classification}

\subsection{Preliminaries}

In the one-class classification task, we are given a set of training samples that are all normal (and contain no anomalies) $x_1,x_2...x_N \in {\cal{X}}_{train}$. The objective is to classify a new sample $x$ as being normal or anomalous. The methods considered here learn a deep representation of a sample parameterized by the neural network function $\phi:{\cal{X}} \rightarrow \mathbb{R}^d$, where $d\in{\mathbb{N}}$ is the feature dimension. In several methods, $\phi$ is initialized by pre-trained weights $\phi_0$, which can be learned either using external datasets (e.g. ImageNet classification) or using self-supervised tasks on the training set. The representation is further tuned on the training data to form the final representation $\phi$. Finally, an anomaly scoring function $s(\phi(x))$ determines the anomaly score of a sample $x$. The binary anomaly classification can be predicted by applying a threshold on $s(x)$. In Sec.~\ref{sec:background:self} and Sec.~\ref{sec:background:pre-trained}, we review the most relevant methods for learning the representation $\phi$. 

\subsection{Self-supervised Objectives for OCC}
\label{sec:background:self}

We review two relevant self-supervised objectives:

\textbf{Center Loss:} This loss uses the simple idea, that features should be learned so that normal data lie within a compact region of feature space, whereas anomalous data lie outside it. As we focus on the OCC setting, there are no examples of anomalies in training. Instead, the center loss encourages the features of the normal samples to lie as near as possible to a predetermined center. Specifically, the center loss for an input sample $x\in {\cal{X}}_{train}$ can be written as follows:
\begin{equation}
\label{eq:center}
    \mathcal{L}_{center}(x) = \|\phi(x) - c\|^2
\end{equation}
This objective suffers from a trivial solution - the features $\phi(x)$ collapse to a singular point $c$ for all samples, normal and anomalous. This is often called "catastrophic collapse". Such a collapsed representation cannot, of course, discriminate between normal and anomalous samples.

\textbf{Contrastive Loss:} Recently, contrastive learning was responsible for much progress in self-supervised representation learning \cite{chen2020simple}. In the contrastive training procedure a mini-batch of size $B$ is randomly sampled and the contrastive prediction task is defined on pairs of augmented examples derived from the mini-batch, resulting in $2B$ data points. For OCC, the contrastive objective simply states that: (i) the angular distance between the features of any positive pair $(x_i, x_{i+B})$ should be small (ii) the distance between the features of normal samples $x_i,x_m\in{{\cal{X}}_{train}},$ where $ i \neq m$, should be large. The typical contrastive loss for a positive pair $(x_i, x_{i+B})$ where $x_i$ and $x_{i+B}$ are augmentations of some $x\in{{\cal{X}}_{train}}$, denoted by $\mathcal{L}_{con}(x_i,x_{i+B})$, is written below:
\begin{equation}
\label{eq:L_contrastive}
     -\log\frac{\textrm{exp}(sim(\phi(x_i), \phi(x_{i+B}))/\tau)}{\sum_{m=1}^{2B} \mathbb{1}{[i \neq m]} \cdot \textrm{exp}(sim(\phi(x_i), \phi(x_m))/\tau)}
\end{equation}
where $\forall{m}\in{[1,2B]}:x_m$ is an augmented view of some $x\in{{\cal{X}}_{train}}$, $\tau$ denotes a temperature hyper-parameter and $sim$ is the cosine similarity. Contrastive methods currently achieve the top performance for OCC without the utilization of externally trained network weights.

\subsection{Initialization with Pre-trained Weights}
\label{sec:background:pre-trained}

Self-supervised representation learning methods have high sample complexity and in many cases do not outperform supervised representation learning methods. It is common practice in deep learning to transfer the weights of classifiers pre-trained on large, somewhat related, labeled datasets to the task of interest. Previous methods used pre-trained weights for anomaly detection \cite{perera2019learning,reiss2021panda}. It was found that fine-tuning the pre-trained weights of $\phi_0$ on the normal data, results in a stronger feature extractor $\phi$. The latest approach, PANDA, used the center loss (Eq.~\ref{eq:center}) for fine-tuning the pre-trained weights. Several attractive properties of methods based on ImageNet pre-trained features were established: (i) they outperform self-supervised anomaly detection methods by a wide margin, without using any labeled examples of anomalies or OE. (ii) they generalize to datasets that are very different from ImageNet including aerial and medical.

As contrastive objectives typically perform better than the center loss, it is natural to assume that replacing PANDA's center loss with the contrastive loss would be advantageous. Unfortunately, the representation collapses immediately and this modification achieves poor OCC results. In Sec.~\ref{sec:method} we will analyze this phenomenon and present an alternative objective that overcomes this issue.

\section{Modifying the Contrastive Loss for Anomaly Detection}
\label{sec:method}
In this section, we introduce our new approach for OCC feature adaptation. In Sec.~\ref{met:contrastive} we analyze the mechanism that prevents standard contrastive objectives from benefiting from pre-trained weights for OCC. In Sec.~\ref{met:approach} we present our new objective function, the mean-shifted contrastive (MSC) loss. In Sec.~\ref{met:msc_explained} we analyze the proposed mean-shifted contrastive loss for OCC transfer learning.

\begin{figure*}[t]
  \centering
    \begin{tabular}{c}
    \includegraphics[scale=1]{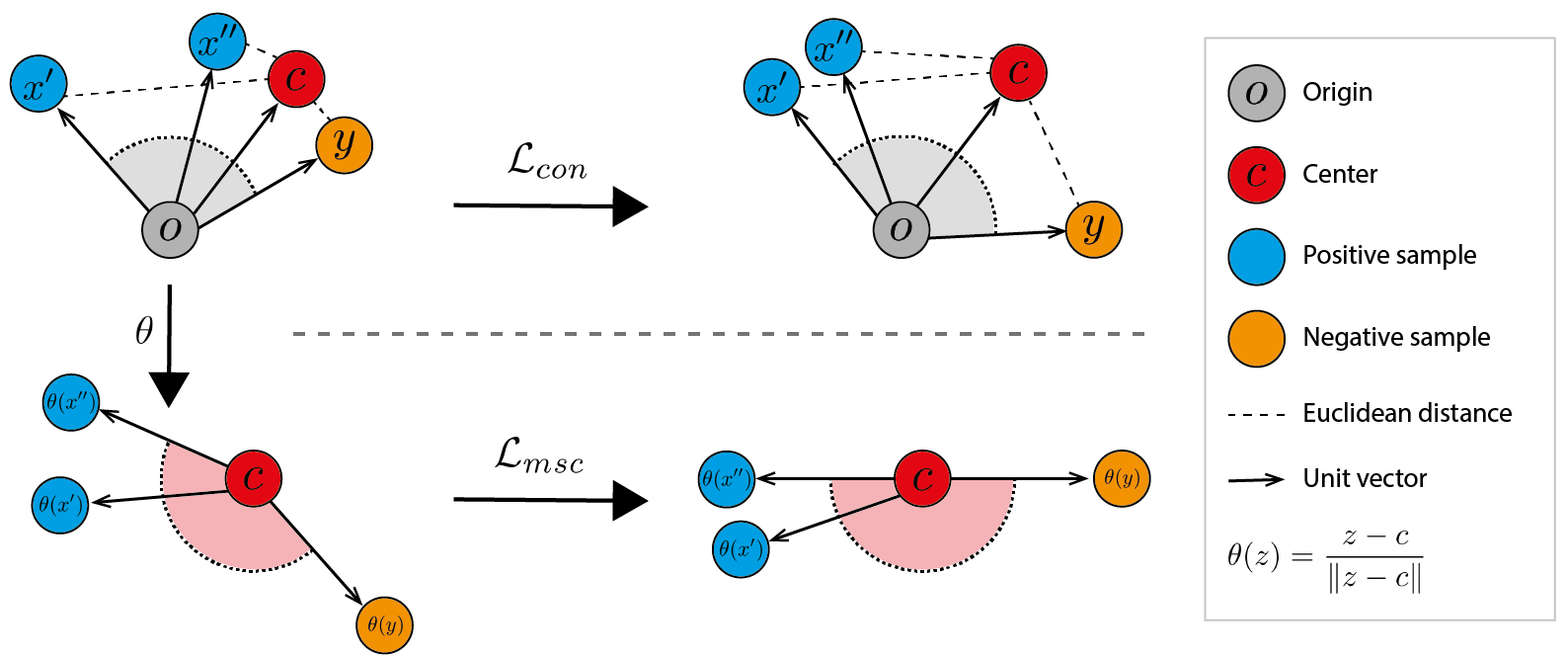} 
    \end{tabular}
    \caption{\textit{Top:} The angular representation in relation to the origin. $\mathcal{L}_{con}$ enlarging the angles between positive and negative samples, thus increasing their Euclidean distance to $c$. \textit{Bottom:} The mean-shifted representation. $\mathcal{L}_{msc}$ does not affect the Euclidean distance between $c$ and the mean-shifted representations while maximizing the angles between the negative pairs.}
    \label{fig:angular_rep}
\end{figure*}

\subsection{Adaptation Failure of The Contrastive Loss}
\label{met:contrastive}

While contrastive methods have achieved state-of-the-art performance on visual recognition tasks, they were not designed for feature adaptation for OCC. Here, we report and analyze a surprising result: when optimizing a contrastive loss for OCC using ImageNet pre-trained features, the representations not only fail to improve but degrade quickly.

To understand this phenomenon, we present in Fig.~\ref{fig:fail_contrastive} plots of two metrics as a function of training epoch: (i) uniformity: the average cosine similarity between the features of pairs of examples in the training set (more uniform = close to zero)  (ii) augmentation distance:  the average cosine similarity between features of train samples and their augmentation (higher generally means better ordering of feature space). \cite{wang2020understanding} showed the contrastive loss optimizes two properties (i) uniform distribution of $\{\phi(x)\}_{x\in{\mathcal{X}_{train}}}$ across the unit sphere. (ii) different augmentations of the same image map to the same representation. We can see that contrastive training improved the uniformity of the distribution but failed to increase the similarity between the features of two views of the same image. Results for other temperatures are in the SM. This shows that contrastive training did not make features more discriminative, suggesting the training objective is not well specified.

We provide an intuitive explanation for the empirical observation. Commonly, the normal data occupy a compact region in the ImageNet pre-trained feature space. When viewed in the spherical coordinate system having its center at the origin, normal images span only a small, bounded region of the sphere. As one of the objectives of contrastive learning is to have features that occupy the entire sphere, the optimization would be focused on changing the features accordingly, putting far less emphasis on improving the features so that they are invariant to augmentations. This is bad for anomaly detection as this uniformity makes anomalies harder to detect (as they become less likely to occupy a sparse region of the feature space). Additionally, such drastic changes in the features cause the loss of the useful properties of the pre-trained feature space. This is counter to the objective of transferring strong auxiliary features.

\subsection{The Mean-Shifted Contrastive Loss for Better Adaptation}
\label{met:approach}

To overcome the limitations of contrastive learning explained above, we propose a simple modification of its objective for OCC feature adaptation. In our modified objective, we compute the angles between the features of images with respect to the \textbf{center} of the normal features rather than the origin (as done in the original contrastive loss). Although this can be seen as a simple shift of the original objective, we will show that it resolves the critical issues highlighted above and allows contrastive learning to benefit from the powerful, pre-trained feature initialization (See Sec.~\ref{met:msc_explained}). We name this new objective, the \textit{Mean-Shifted Contrastive}. 

Let us denote the center of the normalized feature representations of the training set by $c$:
\begin{equation}
\label{eq:features_center}
    c = \mathbb{E}_{x \in {\cal{X}}_{train}} [\frac{\phi_0(x)}{\|\phi_0(x)\|}]
\end{equation}
where $\phi_0$ is the initialized pre-trained model. For each image $x$, we create two different augmentations of the image, denoted $x_i, x_{i+B}$. All the augmented images are first passed through a feature extractor $\phi$. They are then scaled to the unit sphere by $\ell_2$ normalization (see Sec.~\ref{exp:analysis} for the motivation of using $\ell_2$ normalization). We mean-shift each representation, by subtracting the center $c$ from each normalized feature representation. The MSC loss for two augmentations $(x_i, x_{i+B})$ of some image $x\in{{\cal{X}}_{train}}$ from an augmented mini-batch of size $2B$, is defined as follows:
\begin{multline}
\label{eq:L_msc}
    {\mathcal{L}}_{msc}(x_i, x_{i+B}) = -\log (\\\frac{\textrm{exp}(sim(\frac{\phi(x_i)}{\|\phi(x_i)\|} - c, \frac{\phi(x_{i+B})}{\|\phi(x_{i+B})\|} - c)/\tau)}{\sum_{i=1}^{2B} \mathbb{1}_{[i \neq m]} \cdot \textrm{exp}(sim(\frac{\phi(x_i)}{\|\phi(x_i)\|} - c, \frac{\phi(x_m)}{\|\phi(x_m)\|} - c)/\tau)})
\end{multline}
where $\tau$ denotes a temperature hyper-parameter and $sim$ is the cosine similarity. 

\textbf{Anomaly criterion}: To classify a sample as normal or anomalous, we use the cosine similarity from a set of $K$ suitably selected training exemplars $N_k(x)$. The set $N_k(x)$ can be selected by $k$ nearest-neighbors (more accurate) or $k$-means (faster). We compute the cosine similarity between the features of the target image $x$ and the $k$ exemplars $N_k(x)$. The anomaly score is given by:
\begin{equation}
\label{eq:criterion}
    s(x)=\sum_{\phi(y)\in{N_k(x)}} 1 - sim(\phi(x), \phi(y))
\end{equation}
where $sim$ is the cosine similarity. By checking if the anomaly score $s(x)$ is larger than a threshold, we determine if the image $x$ is normal or anomalous. See Sec.~\ref{exp:analysis} for a comparison between different exemplar selection methods.

\begin{figure*}[t]
  \centering
    \begin{tabular}{c}
    \includegraphics[scale=1.05]{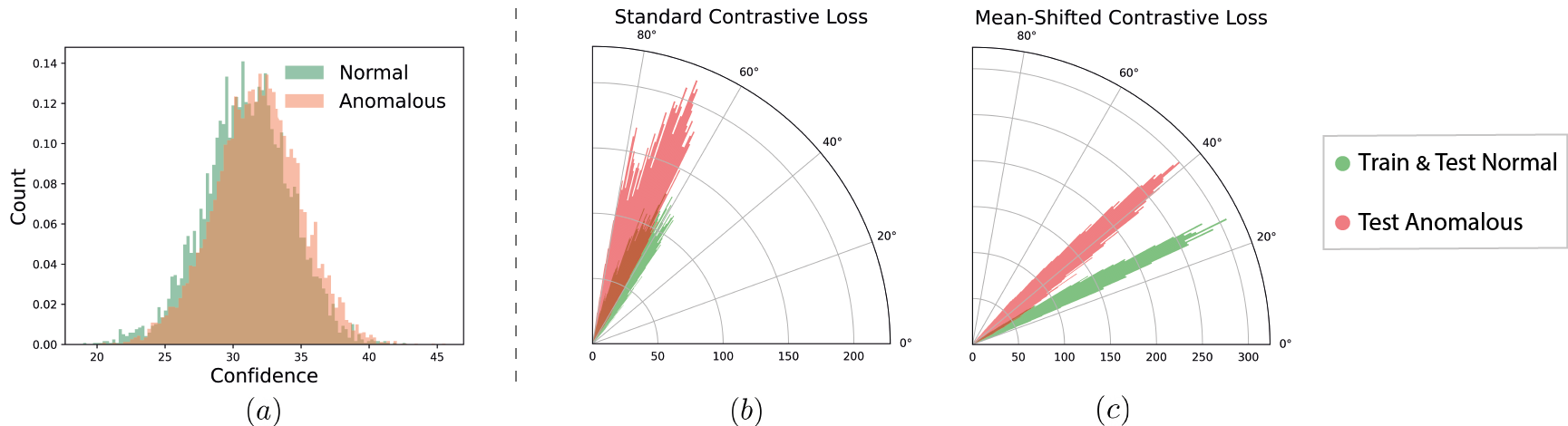} 
    \end{tabular}
    \caption{CIFAR-10 "Bird" class. \textit{Left: (a)} Confidence histogram. The $\ell_2$ norm confidence of the extracted features derived by $\phi$ does not differentiate between normal and anomalous samples. \textit{Right: }Histograms of the angular distance between images to the center measured around the origin for \textit{(b)} Standard Contrastive objective \textit{(c)} The Mean-shifted contrastive objective.}
     \label{fig:polar}

\end{figure*}

\subsection{Understanding the Mean-Shifted Loss}
\label{met:msc_explained}

\textbf{Uniformity:} Optimizing pre-trained weights with the standard contrastive loss focuses on optimizing uniformity around the origin-centered sphere but hurts feature semantic similarity (Sec.~\ref{met:contrastive}). The MSC loss proposes a simple but very effective solution - evaluating uniformity in the coordinate frame around the data center. As the features in this frame are already roughly uniform, the optimization focuses on improving their semantic similarity. As shown in Fig.~\ref{fig:fail_contrastive}, the features are uniform right from initialization with our objective (low cosine similarity between normal examples). Thus, the optimization can focus on improving the features.

\textbf{Compactness around center:} The standard contrastive loss maximizes the angles between representations of negative pairs even when they are both normal training images. By maximizing these angles, the distance to the center increases as well, as illustrated in Fig.~\ref{fig:angular_rep} (top). This behavior is in contrast to the optimization of the center loss (Eq.~\ref{eq:center}), which learns representations by minimizing the Euclidean distance between normal representations and the center. \cite{reiss2021panda} showed that optimizing the center loss results in high anomaly detection performance. Our proposed loss does not suffer from this issue. Instead of measuring the angular distance between samples in relation to the origin, we measure the angular distance in relation to the center of the normal features. As can be seen in Fig.~\ref{fig:angular_rep} (bottom), our proposed loss maximizes the angles between the negative pairs while preserving their distance to the center.

In Fig.~\ref{fig:polar}.b-c we present histograms of the angular distance between images to the center measured around the origin with (i) standard contrastive loss and (ii) our MSC loss. The distributions of normal and anomalous features overlap in standard contrastive, but not in our MSC loss.

\section{Experiments}
\label{sec:exp}

In this section, we extensively evaluate our method. In Sec.~\ref{exp:main}, we report our OCC results with a comparison to previous works on the standard benchmark datasets. In Sec.~\ref{exp:analysis} we further analyze our objective and we present an ablation study. Building upon the framework suggested in \cite{reiss2021panda}, we use ResNet152 pre-trained on ImageNet classification task as $\phi_0$ (unless specified otherwise), and adding an additional final $\ell_2$ normalization layer - this is our initialized feature extractor $\phi$.  By default, we fine-tune our model with $\mathcal{L}_{msc}$ (as in Eq.~\ref{eq:L_msc}). For inference, we use the criterion described in Sec.~\ref{met:approach}. We adopt the ROC-AUC metric as the detection performance score. Full training and implementation details are in the Supplementary Material (SM).

\subsection{Main Results}
\label{exp:main}
We evaluated our approach on a wide range of anomaly detection benchmarks. Following \cite{golan2018deep,hendrycks2019using}, we run our experiments on commonly used datasets:  CIFAR-10 \cite{krizhevsky2009learning}, CIFAR-100 coarse-grained version that consists of 20 classes \cite{krizhevsky2009learning}, and CatsVsDogs \cite{elson2007asirra}. Following standard protocol, multi-class datasets are converted to anomaly detection by setting a class as normal and all other classes as anomalies. This is performed for all classes, in practice turning a single dataset with $C$ classes into $C$ datasets. We compare our approach with the top current self-supervised and pre-trained feature adaptation methods \cite{ruff2018deep,hendrycks2019using,tack2020csi,sohn2020learning,reiss2021panda}. Full dataset descriptions and baselines are in the SM.

Tab.~\ref{tab:datasets} shows that our proposed approach surpasses the previous state-of-the-art on the common OCC benchmarks. This establishes the superiority of our approach, resulted from our new objective, over previous self-supervised and pre-trained methods. Full class-wise results are in the SM.

\begin{table*}[t]
  \centering
  \begin{tabular}{lccccccc}
    \toprule
	& \multicolumn{4}{c}{Self-supervised} 	&	\multicolumn{3}{c}{Pre-trained} 	\\
\cmidrule(r){1-1}
\cmidrule(r){2-5}
\cmidrule(r){6-8}
Architecture	&	LeNet 	& 	\multicolumn{4}{c}{ResNet18} & \multicolumn{2}{c}{ResNet152} \\

	\cmidrule(r){1-1}
    \cmidrule(r){2-2}
    \cmidrule(r){3-6}    
    \cmidrule(r){7-8}
Method    	&	DeepSVDD 	&	MRot 	&	DROC     &   CSI &  ${\mathcal{L}}_{msc}$ & PANDA & ${\mathcal{L}}_{msc}$ \\
    \cmidrule(r){1-1}
    \cmidrule(r){2-2}
    \cmidrule(r){3-6}    
    \cmidrule(r){7-8}
CIFAR-10
	&	64.8	&	90.1	&	92.5	&	94.3	&	\textbf{94.8}  & 96.2 &   \textbf{97.2}\\
CIFAR-100
	&	67.0	&	80.1	&	86.5	&	89.6	&	\textbf{94.4}  & 94.1 &   \textbf{96.4}	\\
CatsVsDogs	&	50.5	&	86.0	&	89.6	&	86.3	&	\textbf{98.4}    & 97.3 & \textbf{99.3}    \\
    \bottomrule
  \end{tabular}
\caption{Anomaly detection performance (mean ROC-AUC\%). Best in bold.}
\label{tab:datasets}
\end{table*}

\begin{table*}[t]
  \centering
  \begin{tabular}{lcccc}
    \toprule
Dataset & CSI  & PANDA & ${\mathcal{L}}_{msc}$ (ResNet18) & ${\mathcal{L}}_{msc}$ (ResNet152) \\
\midrule
DIOR &  78.5 & 94.3 & 97.5 & \textbf{97.7}  \\
MvTec & 63.6 & 86.5 & 85.0 & \textbf{87.2} \\
CIFAR-100 (Fine-grained)    &	90.1	&	97.1	&	92.0	& \textbf{97.6}	\\		
CIFAR-10 (200 samples)    &	81.3	&	95.4	&	93.1	& \textbf{96.5}	\\
CIFAR-10 (500 samples) &	88.1	&	95.6	&	93.8	& \textbf{96.7} \\
    \bottomrule
  \end{tabular}
    \caption{Anomaly detection accuracy (mean ROC-AUC\%) on small dataset. Best in bold.}
    \label{tab:norm_dataset}
\end{table*}

\subsection{Further Analysis \& Ablation Study}
\label{exp:analysis}

\textbf{Small datasets.} To demonstrate different challenges in image anomaly detection, we further extend our results on small datasets following the standard protocol. We tested our method on: MVTec \cite{bergmann2019mvtec} and DIOR \cite{li2020object} and the fine-grained version of CIFAR-100 (100 classes). Furthermore, we used the CIFAR-10 dataset with different amounts of training data. In Tab.~\ref{tab:norm_dataset} we present a comparison between (i) the top self-supervised contrastive-learning based method - CSI (ii) the top OCC feature adaptation method - PANDA (iii) our method. We see that the self-supervised method does not perform well on such small datasets, whereas our method achieves very strong performance. The reason for the poor performance of self-supervised methods on small datasets is that they only see the small dataset for training, and cannot learn strong features using such a small sample size. This is particularly severe for contrastive methods (but is also the case for all other self-supervised methods). As pre-trained methods transfer features from external datasets, they perform better.

\begin{figure*}[t]
  \centering
    \begin{tabular}{c}
    \includegraphics[scale=1.05]{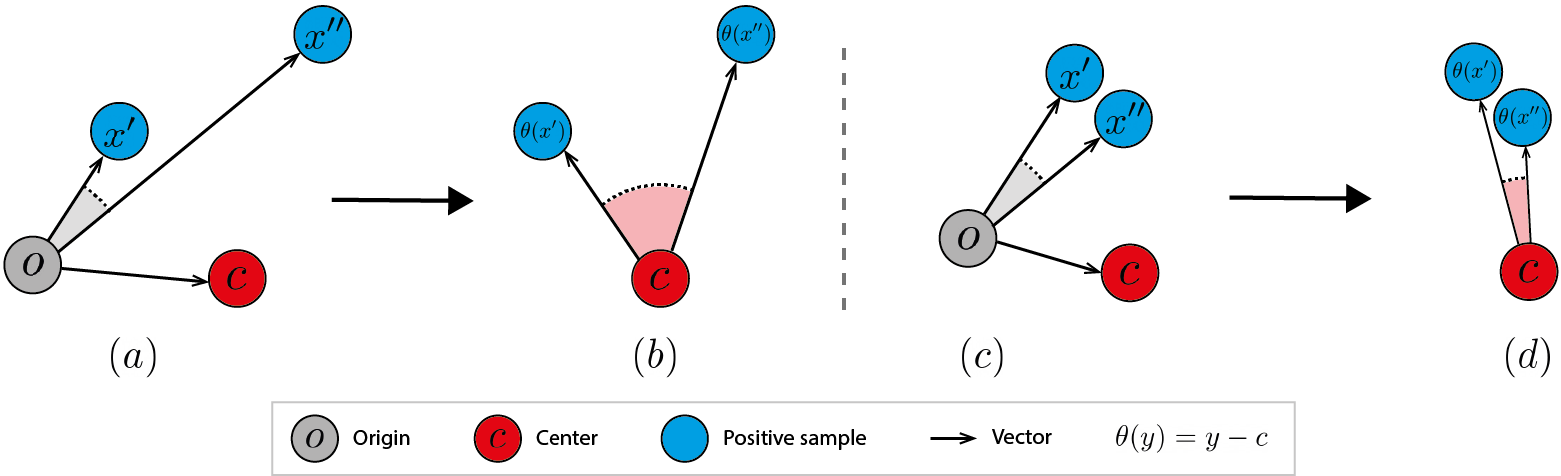} 
    \end{tabular}
    \caption{Sensitivity of ${\mathcal{L}}_{msc}$ to class confidence. \textit{(a)} The angular representation in relation to the origin without confidence normalization. \textit{(b)} The mean-shifted representation enlarges the angle between the positive samples. \textit{(c)} The angular representation after confidence normalization. \textit{(d)} The angle between the positive samples is  approximately preserved after mean-shifting.} \label{fig:angular_explained}
\end{figure*}

\textbf{The Angular Representation.} Our initial feature extractor $\phi_0$ is pre-trained on a classification task (ImageNet classification). To obtain class probabilities, the features $\phi_0(x)$ are subsequently multiplied by classifier matrix $C$ and passed through a softmax layer. The logits are therefore given by $C \phi_0(x)$. As softmax is a monotonic function, scaling of the logits does not change the order of probabilities. However, scaling does determine the degree of confidence in the decision. We propose to disambiguate the representation $\phi_0(x)$ into two components: (i) the semantic class $\frac{\phi_0(x)}{\|\phi_0(x)\|}$, and (ii) the confidence $\|\phi_0(x)\|$. The confidence acts as a per-sample temperature that determines how confident the discrimination between the classes is. A thorough investigation that we conducted, showed that the confidence of an ImageNet pre-trained feature representation did not help the anomaly detection performance. In Fig.~\ref{fig:polar}.a, we compare the histogram of confidence values between the normal and anomalous values on a particular class of the CIFAR-10 dataset. We observe that confidence does not discriminate between normal and anomalous images in this dataset. In Fig.~\ref{fig:angular_explained} we demonstrate the sensitivity of the mean-shifted representation to class confidence. This emphasizes the importance of confidence normalization for MSC optimization.

We thus propose to use the angular center loss. The angular center loss encourages the angular distance between each sample and the center to be minimal. This contrasts with the standard center loss (used by PANDA and DeepSVDD), which uses the Euclidean distance. Although a simple change, the angular center loss achieves much better results than the regular center loss (see Tab.~\ref{tab:ablation}).
\begin{equation}
\label{eq:L_sph_center}
   {\mathcal{L}}_{ang} = -\phi(x) \cdot c
\end{equation}

\begin{table}[t]
  \centering
  \begin{tabular}{cccccc}

\multicolumn{2}{c}{DN2}	&	\multicolumn{2}{c}{PANDA} 	&	${\mathcal{L}}_{msc}$	&	${\mathcal{L}}_{msc} + {\mathcal{L}}_{ang}$		\\
\cmidrule(r){1-2}\cmidrule(r){3-4}\cmidrule(r){5-5}\cmidrule(r){6-6}
Raw & Ang. & ${\mathcal{L}}_{center}$ & ${\mathcal{L}}_{ang}$ & & \\
\midrule
92.5	&	95.8	&	96.2	& 96.8  &	 97.2	&	\textbf{97.5}		\\											
    \bottomrule
  \end{tabular}
    \caption{Training objection ablation study (CIFAR-10, mean ROC-AUC\%).}
    \label{tab:ablation}
\end{table}

\textbf{Training objective.} An ablation of the objectives and of DN2 ($k$NN on unadapted ImageNet pre-trained ResNet features) is presented in Tab.~\ref{tab:ablation}. Note that both the confidence-invariant form of DN2 and PANDA outperform their Euclidean versions. We further notice that the MSC loss outperforms the rest, and combining it with the angular center loss results in further improvements.

\textbf{Optimization from scratch.} The mean-shifted objective assumes that the relative distance to the center of the features is correlated with high detection performance. When initializing the center as a random Gaussian vector we lose this strong prior, as a result, the detection capabilities are drastically degraded. Therefore when training a model from scratch without any strong initialization that comes from a pre-trained model, our objective does not improve over standard contrastive losses. The MSC loss is therefore a directed contribution to anomaly detection from pre-trained features.

\begin{table}[t]
  \centering
  \begin{tabular}{lcccc}
    \toprule

Method 	&	ResNet &  EfficientNet	& DenseNet & ViT \\
\midrule
DN2
	&	92.5	& 89.3 &	 85.6 & 95.7	\\
PANDA
	&	96.2	&	95.3 & 82.4 & 95.8	\\	
${\mathcal{L}}_{msc}$
	&	\textbf{97.2}	& \textbf{97.0}	 & \textbf{95.7} & \textbf{98.6}	\\	
    \bottomrule
  \end{tabular}
    \caption{Performance gains with different network architectures (CIFAR-10, mean ROC-AUC\%). Best in bold.}
    \label{tab:arc}
\end{table}

\textbf{Do ImageNet pre-trained features extend to distant domains?} Our results on DIOR and MVTec which are significantly different from ImageNet provide such evidence. This was also extensively established by \cite{reiss2021panda}. 

\textbf{Catastrophic collapse \& Early Stopping.} Like other OCC pre-trained feature adaptation methods (e.g. PANDA), our method suffers from catastrophic collapse after a very large number of training epochs. However, our method is less sensitive than PANDA, as we dominate PANDA at any point in the curve and collapse much more slowly. See SM.

\textbf{Why do self-supervised OCC models not suffer from catastrophic collapse?} Pre-trained methods start from highly discriminative features and can therefore lose accuracy whereas self-supervised features start from random features and therefore have nothing to forget.

\textbf{Rotation prediction methods do not benefit from pre-trained features.} The best performing self-supervised approaches \cite{tack2020csi,sohn2020learning} use a combination of contrastive and rotation prediction objectives. Although initially counter-intuitive, AD by rotation prediction does not benefit from pre-trained features. The reason is that features that generalize better (e.g. ImageNet pre-trained), achieve better performance on rotation prediction for \textit{both} normal and anomalous data. Pre-training, therefore, decreases the gap between the performance of normal and anomalous images on rotation prediction in comparison to randomly-initialized networks. As this gap is used for discriminating between normal and anomalous samples, its decrease leads to worse anomaly detection performance. Specifically, we found that CSI with ImageNet pre-trained features achieves $89.5\%$ on CIFAR-10 compared to the standard version which results in $94.3\%$.

\textbf{Self-supervised methods do not benefit from large architectures.} Pre-trained models can use large deep networks, a quality that self-supervised OCC methods lack as OCC benchmarks are not large. We tested this by evaluating CSI with different ResNet backbone sizes (ResNet18, 50, 152). The CSI results were the same for all backbones sizes $94.3\%$ ROC-AUC on CIFAR-10. This contrasts with pre-trained feature adaptation in our method which benefits from bigger backbones and also outperforms CSI on the same-sized backbone (ResNet18).

\begin{table}[t]
  \centering
  \begin{tabular}{ccccccc}
    \toprule
 \multicolumn{2}{c}{DN2}	&	\multicolumn{2}{c}{PANDA} 	&	\multicolumn{2}{c}{SSL}	&	Ours \\
 \cmidrule(r){1-2} \cmidrule(r){3-4} \cmidrule(r){5-6} \cmidrule(r){7-7}
Raw & Ang. & ${\mathcal{L}}_{center}$ & ${\mathcal{L}}_{ang}$ & MRot & CSI & ${\mathcal{L}}_{msc}$ \\
\midrule
 76.2	&	80.4	&	78.5	& 78.0  &	 76.7	& 79.0 &	\textbf{85.3}	\\											
    \bottomrule
  \end{tabular}
\caption{Multi-Class AD (CIFAR-10, mean ROC-AUC\%).}
\label{tab:multi-modal}
\end{table}

\textbf{Performance on different network architectures.} In Tab.~\ref{tab:arc} we provide the CIFAR-10 ROC-AUC\% results of DN2, PANDA and our method on leading CNN and ViT architectures pre-trained on ImageNet. PANDA is sensitive to the choice of architecture and did not improve results on ViT and collapsed on DenseNet. Our MSC loss generalizes across architectures and results in significant performance gains, including $98.6 \%$ ROC-AUC on CIFAR-10.

\textbf{Multi-Class Anomaly Detection.} We evaluate the setting introduced by \cite{ahmed2020detecting} where the normal data contains multiple semantic classes. Note that \textit{no class labels are provided} for the normal data. This setting is more challenging than the single-class setting as the normal data has a multi-modal distribution. For each experiment, we denoted a single CIFAR-10 class as anomalous and other CIFAR-10 classes as normal. We report the mean ROC-AUC\% over the 10 experiments in Tab.~\ref{tab:multi-modal}. PANDA provides no improvement over DN2 (with cosine distance) as the data is no longer uni-modal and therefore not compact. In contrast, our MSC loss does not rely on the uni-modal assumption to the same extent and produces better results than previous pre-trained and self-supervised methods.

\begin{table}[t]
  \centering
  \begin{tabular}{ccccc}
    \toprule
$k=1$ & $k=5$ & $k=10$ & $k=100$ & Full train set \\
\midrule
94.2 & 95.8 & 96.1 & 97.0 & 97.2  \\
    \bottomrule
  \end{tabular}
\caption{CIFAR-10 OCC with $k$-means (mean ROC-AUC\%)}
\label{tab:kmeans}
\end{table}

\textbf{Pre-training and Outlier Exposure (OE).} AD methods employ different levels of supervision. The most extensive supervision is used by OE \cite{deecke2021transfer,ruff2020rethinking}, which requires a large external dataset at training time, and requires the dataset to be from a similar domain to the anomalies. OE methods fail when the OE is more similar to the normal than the anomalous data have different distributions, as anomalous data are more likely to be classified as normal than OE. This can be overcome by using supervision to select an OE dataset that simulates the anomalies, but this contradicts the objective of anomaly detection, the detection of any type of anomaly. Pre-training, like OE, is also achieved through an external labeled dataset, but differently from OE, the labels are not required to be related to the anomalies and the external dataset is only required once - at the pre-training stage and is never used again.

\textbf{Detection scoring functions.} $k$NN has well-established approximations that mitigate its inference time complexity. A simple, but effective solution is reducing the set of gallery samples via $k$-means. In Tab.~\ref{tab:kmeans} we present a comparison of the performance of our method and its $k$-means approximations with the features of the normal training images compressed using different numbers of means ($k$). We use the sum of the distances to the nearest neighbor means as the anomaly score. We can see that significant inference time improvement can be achieved for a small loss in accuracy.

\section{Conclusion}
This paper investigated feature adaptation methods for anomaly detection. We first analyzed the standard contrastive loss and found that it provides a poor initialization for OCC feature adaptation. We then introduced the Mean-Shifted Contrastive loss, which overcomes the limitations of the standard contrastive loss. Extensive experiments verified the outstanding performance of our method.

\section*{Acknowledgments}
We are grateful to Shira Bar-On for making our lovely figures. Tal Reiss was funded by grants from the Israeli Cyber Directorate, the Israeli Council for Higher Education, and a Facebook Research Award.

\bibliography{aaai23}

\begin{thebibliography}{31}
\providecommand{\natexlab}[1]{#1}

\bibitem[{Ahmed and Courville(2020)}]{ahmed2020detecting}
Ahmed, F.; and Courville, A. 2020.
\newblock Detecting semantic anomalies.
\newblock In \emph{Proceedings of the AAAI Conference on Artificial
  Intelligence}, volume~34, 3154--3162.

\bibitem[{Bergman and Hoshen(2020)}]{bergman2020classification}
Bergman, L.; and Hoshen, Y. 2020.
\newblock Classification-Based Anomaly Detection for General Data.
\newblock In \emph{ICLR}.

\bibitem[{Bergmann et~al.(2019)Bergmann, Fauser, Sattlegger, and
  Steger}]{bergmann2019mvtec}
Bergmann, P.; Fauser, M.; Sattlegger, D.; and Steger, C. 2019.
\newblock MVTec AD--A Comprehensive Real-World Dataset for Unsupervised Anomaly
  Detection.
\newblock In \emph{Proceedings of the IEEE Conference on Computer Vision and
  Pattern Recognition}, 9592--9600.

\bibitem[{Chen et~al.(2020{\natexlab{a}})Chen, Kornblith, Norouzi, and
  Hinton}]{chen2020simple}
Chen, T.; Kornblith, S.; Norouzi, M.; and Hinton, G. 2020{\natexlab{a}}.
\newblock A simple framework for contrastive learning of visual
  representations.
\newblock \emph{arXiv preprint arXiv:2002.05709}.

\bibitem[{Chen et~al.(2020{\natexlab{b}})Chen, Fan, Girshick, and
  He}]{chen2020improved}
Chen, X.; Fan, H.; Girshick, R.; and He, K. 2020{\natexlab{b}}.
\newblock Improved baselines with momentum contrastive learning.
\newblock \emph{arXiv preprint arXiv:2003.04297}.

\bibitem[{Deecke et~al.(2021)Deecke, Ruff, Vandermeulen, and
  Bilen}]{deecke2021transfer}
Deecke, L.; Ruff, L.; Vandermeulen, R.~A.; and Bilen, H. 2021.
\newblock Transfer-based semantic anomaly detection.
\newblock In \emph{International Conference on Machine Learning}, 2546--2558.
  PMLR.

\bibitem[{Elson et~al.(2007)Elson, Douceur, Howell, and Saul}]{elson2007asirra}
Elson, J.; Douceur, J.~R.; Howell, J.; and Saul, J. 2007.
\newblock Asirra: a CAPTCHA that exploits interest-aligned manual image
  categorization.
\newblock In \emph{ACM Conference on Computer and Communications Security},
  volume~7, 366--374.

\bibitem[{Eskin et~al.(2002)Eskin, Arnold, Prerau, Portnoy, and
  Stolfo}]{eskin2002geometric}
Eskin, E.; Arnold, A.; Prerau, M.; Portnoy, L.; and Stolfo, S. 2002.
\newblock A geometric framework for unsupervised anomaly detection.
\newblock In \emph{Applications of data mining in computer security}, 77--101.
  Springer.

\bibitem[{Gidaris, Singh, and Komodakis(2018)}]{gidaris2018unsupervised}
Gidaris, S.; Singh, P.; and Komodakis, N. 2018.
\newblock Unsupervised representation learning by predicting image rotations.
\newblock \emph{arXiv preprint arXiv:1803.07728}.

\bibitem[{Glodek, Schels, and Schwenker(2013)}]{glodek2013ensemble}
Glodek, M.; Schels, M.; and Schwenker, F. 2013.
\newblock Ensemble Gaussian mixture models for probability density estimation.
\newblock \emph{Computational Statistics}, 28(1): 127--138.

\bibitem[{Golan and El-Yaniv(2018)}]{golan2018deep}
Golan, I.; and El-Yaniv, R. 2018.
\newblock Deep Anomaly Detection Using Geometric Transformations.
\newblock In \emph{NeurIPS}.

\bibitem[{Hendrycks et~al.(2019)Hendrycks, Mazeika, Kadavath, and
  Song}]{hendrycks2019using}
Hendrycks, D.; Mazeika, M.; Kadavath, S.; and Song, D. 2019.
\newblock Using self-supervised learning can improve model robustness and
  uncertainty.
\newblock In \emph{NeurIPS}.

\bibitem[{Jolliffe(2011)}]{jolliffe2011principal}
Jolliffe, I. 2011.
\newblock \emph{Principal component analysis}.
\newblock Springer.

\bibitem[{Kirkpatrick et~al.(2017)Kirkpatrick, Pascanu, Rabinowitz, Veness,
  Desjardins, Rusu, Milan, Quan, Ramalho, Grabska-Barwinska
  et~al.}]{kirkpatrick2017overcoming}
Kirkpatrick, J.; Pascanu, R.; Rabinowitz, N.; Veness, J.; Desjardins, G.; Rusu,
  A.~A.; Milan, K.; Quan, J.; Ramalho, T.; Grabska-Barwinska, A.; et~al. 2017.
\newblock Overcoming catastrophic forgetting in neural networks.
\newblock \emph{Proceedings of the national academy of sciences}, 114(13):
  3521--3526.

\bibitem[{Krizhevsky, Hinton et~al.(2009)}]{krizhevsky2009learning}
Krizhevsky, A.; Hinton, G.; et~al. 2009.
\newblock Learning multiple layers of features from tiny images.
\newblock \emph{Citeseer}.

\bibitem[{Larsson, Maire, and Shakhnarovich(2016)}]{larsson2016learning}
Larsson, G.; Maire, M.; and Shakhnarovich, G. 2016.
\newblock Learning representations for automatic colorization.
\newblock In \emph{ECCV}.

\bibitem[{Latecki, Lazarevic, and Pokrajac(2007)}]{latecki2007outlier}
Latecki, L.~J.; Lazarevic, A.; and Pokrajac, D. 2007.
\newblock Outlier detection with kernel density functions.
\newblock In \emph{International Workshop on Machine Learning and Data Mining
  in Pattern Recognition}, 61--75. Springer.

\bibitem[{Li et~al.(2020)Li, Wan, Cheng, Meng, and Han}]{li2020object}
Li, K.; Wan, G.; Cheng, G.; Meng, L.; and Han, J. 2020.
\newblock Object detection in optical remote sensing images: A survey and a new
  benchmark.
\newblock \emph{ISPRS Journal of Photogrammetry and Remote Sensing}, 159:
  296--307.

\bibitem[{Mathieu, Couprie, and LeCun(2016)}]{mathieu2015deep}
Mathieu, M.; Couprie, C.; and LeCun, Y. 2016.
\newblock Deep multi-scale video prediction beyond mean square error.
\newblock \emph{ICLR}.

\bibitem[{Noroozi and Favaro(2016)}]{noroozi2016unsupervised}
Noroozi, M.; and Favaro, P. 2016.
\newblock Unsupervised learning of visual representations by solving jigsaw
  puzzles.
\newblock In \emph{ECCV}.

\bibitem[{Perera and Patel(2019)}]{perera2019learning}
Perera, P.; and Patel, V.~M. 2019.
\newblock Learning deep features for one-class classification.
\newblock \emph{IEEE Transactions on Image Processing}, 28(11): 5450--5463.

\bibitem[{Reiss et~al.(2021)Reiss, Cohen, Bergman, and Hoshen}]{reiss2021panda}
Reiss, T.; Cohen, N.; Bergman, L.; and Hoshen, Y. 2021.
\newblock PANDA: Adapting Pretrained Features for Anomaly Detection and
  Segmentation.
\newblock In \emph{Proceedings of the IEEE/CVF Conference on Computer Vision
  and Pattern Recognition}, 2806--2814.

\bibitem[{Ruff et~al.(2018)Ruff, Gornitz, Deecke, Siddiqui, Vandermeulen,
  Binder, M{\"u}ller, and Kloft}]{ruff2018deep}
Ruff, L.; Gornitz, N.; Deecke, L.; Siddiqui, S.~A.; Vandermeulen, R.; Binder,
  A.; M{\"u}ller, E.; and Kloft, M. 2018.
\newblock Deep one-class classification.
\newblock In \emph{ICML}.

\bibitem[{Ruff et~al.(2020)Ruff, Vandermeulen, Franks, M{\"u}ller, and
  Kloft}]{ruff2020rethinking}
Ruff, L.; Vandermeulen, R.~A.; Franks, B.~J.; M{\"u}ller, K.-R.; and Kloft, M.
  2020.
\newblock Rethinking assumptions in deep anomaly detection.
\newblock \emph{arXiv preprint arXiv:2006.00339}.

\bibitem[{Scholkopf et~al.(2000)Scholkopf, Williamson, Smola, Shawe-Taylor, and
  Platt}]{scholkopf2000support}
Scholkopf, B.; Williamson, R.~C.; Smola, A.~J.; Shawe-Taylor, J.; and Platt,
  J.~C. 2000.
\newblock Support vector method for novelty detection.
\newblock In \emph{NIPS}.

\bibitem[{Sohn et~al.(2020)Sohn, Li, Yoon, Jin, and Pfister}]{sohn2020learning}
Sohn, K.; Li, C.-L.; Yoon, J.; Jin, M.; and Pfister, T. 2020.
\newblock Learning and Evaluating Representations for Deep One-class
  Classification.
\newblock \emph{arXiv preprint arXiv:2011.02578}.

\bibitem[{Tack et~al.(2020)Tack, Mo, Jeong, and Shin}]{tack2020csi}
Tack, J.; Mo, S.; Jeong, J.; and Shin, J. 2020.
\newblock Csi: Novelty detection via contrastive learning on distributionally
  shifted instances.
\newblock \emph{NeurIPS}.

\bibitem[{Tax and Duin(2004)}]{tax2004support}
Tax, D.~M.; and Duin, R.~P. 2004.
\newblock Support vector data description.
\newblock \emph{Machine learning}.

\bibitem[{Wang and Liu(2021)}]{wang2021understanding}
Wang, F.; and Liu, H. 2021.
\newblock Understanding the behaviour of contrastive loss.
\newblock In \emph{Proceedings of the IEEE/CVF Conference on Computer Vision
  and Pattern Recognition}, 2495--2504.

\bibitem[{Wang and Isola(2020)}]{wang2020understanding}
Wang, T.; and Isola, P. 2020.
\newblock Understanding contrastive representation learning through alignment
  and uniformity on the hypersphere.
\newblock In \emph{International Conference on Machine Learning}, 9929--9939.
  PMLR.

\bibitem[{Zhang, Isola, and Efros(2016)}]{zhang2016colorful}
Zhang, R.; Isola, P.; and Efros, A.~A. 2016.
\newblock Colorful image colorization.
\newblock In \emph{ECCV}.

\end{thebibliography}

\newpage
\clearpage

\appendix

\section{Experimental details}

\label{appendix:exp_details}

\subsection{Dataset Descriptions}
\label{appendix:datasets}
\textbf{Standard datasets:} We evaluate our method on a set of commonly used datasets: \textit{CIFAR-10} \cite{krizhevsky2009learning}:  Consists of RGB images of 10  object classes. \textit{CIFAR-100} \cite{krizhevsky2009learning}:  We use the coarse-grained version that consists of 20 classes. \textit{DogsVsCats}: High resolution color images of two classes: cats and dogs. The data were extracted from the ASIRRA dataset \cite{elson2007asirra}, we split each class to the first $10,000$ images as train and the last 2,500 as test.

\textbf{Small datasets:} To further extend our results, we compared the methods on a number of small datasets from different domains. \textit{MvTec} \cite{bergmann2019mvtec}: This dataset contains 15 different industrial products, with normal images of proper products for train and $1-9$ types of manufacturing errors as anomalies. The anomalies in MvTec are in-class i.e. the anomalous images come from the same class of normal images with subtle variations. \textit{DIOR} \cite{li2020object}: We pre-processed the DIOR aerial image dataset by taking the segmented object in classes that have more than $50$ images with size larger than $120\times120$ pixels. 

\subsection{Baselines}
\label{appendix:baselines}
\textit{DROC} \cite{sohn2020learning}: We used the numbers reported in the paper.

For the evaluation of the other competing method, we trained using the official repositories of their authors and make an effort to select the best configurations available.

\textit{DeepSVDD} \cite{ruff2018deep}: We resize all the images to $32 \times 32$ pixels and use the official PyTorch implementation with the CIFAR-10 configuration.

\textit{MRot} \cite{hendrycks2019using}: An improved version of the original RotNet approach. For high-resolution images we used the current GitHub implementation. For low resolution images, we modified the code to the architecture described in the paper, replicating the numbers in the paper on CIFAR-10.

\textit{CSI} \cite{tack2020csi}, \textit{PANDA} \cite{reiss2021panda}: We run the code and used the exact protocol as described in the official repositories.

\subsection{Implementation details}
\label{appendix:imp_details}
For the ResNet152 experiments we fine-tune the two last blocks of an ImageNet pre-trained ResNet152 with an additional $\ell_2$ normalization layer for 25 epochs. For the ResNet18 experiments we fine-tune the full backbone of an ImageNet pre-trained ResNet18 with an additional $\ell_2$ normalization layer for 20 epochs. In both settings we minimize $\mathcal{L}_{msc}$ where the temperature $\tau$ is set as 0.25.  We use SGD optimizer with weight decay of $w = 5\cdot10^{-5}$, and no momentum. The size of the mini-batches is set to be $64$. We adopt the data augmentation module proposed by \cite{chen2020improved}; we sequentially apply a $224\times224$-pixel crop from a randomly resized image, random color jittering, random grayscale conversion, random Gaussian blur and random horizontal flip. Finally, for anomaly scoring we use kNN with $k=2$ nearest neighbours. 

\subsection{Training Resources}
Training each dataset class presented in this paper takes approximately 3 hours on a single NVIDIA RTX-2080 TI.

\subsection{Per-class results}
\label{appendix:per_class}
In Tab.~\ref{tab:cifar-10}, Tab.~\ref{tab:cifar-100}, Tab.\ref{tab:catsdogs} we present the per-class results of CIFAR-10, CIFAR-100, CatsVsDogs respectively.

\section{Catastrophic Collapse}
\label{appendix:collapse}
\begin{figure*}[ht]

  \centering
    \begin{tabular}{c}
\includegraphics[scale=0.75]{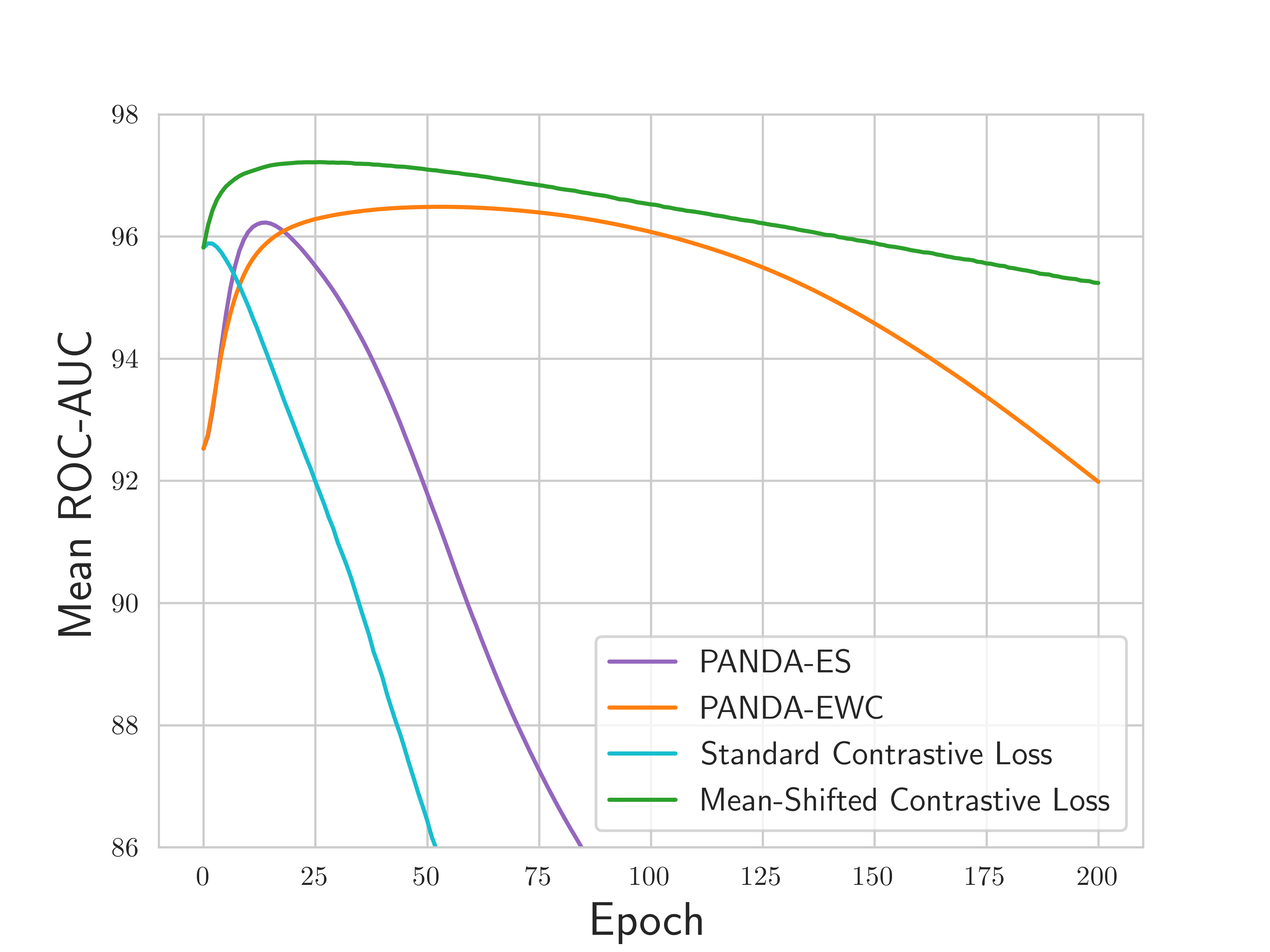}
    \end{tabular}
    \caption{CIFAR-10 Mean ROC-AUC\%. Catastrophic collapse of various objective functions.}      \label{appendix:fig}
\end{figure*}

In Fig.~\ref{appendix:fig}, we evaluated the collapse of different training objectives averaged on all CIFAR-10 classes. We notice that the contrastive loss is unsuitable for OCC feature adaptation as it results in very fast catastrophic collapse. PANDA-ES (early-stopping) results in initial improvement in accuracy, but after few epochs the features degrade and become uninformative. PANDA-EWC postpones the collapse, but does not prevent it. Finally, we see that the mean-shifted contrastive loss dominates PANDA at any point in the curve and collapses much more slowly. We find that early stopping after 25 iterations typically gets very close to the optimal accuracy. 

\section{The Temperature parameter and Uniformity}
\label{appendix:tau}
\begin{figure*}[ht]
  \centering
    \begin{tabular}{c}
\includegraphics[scale=1.45]{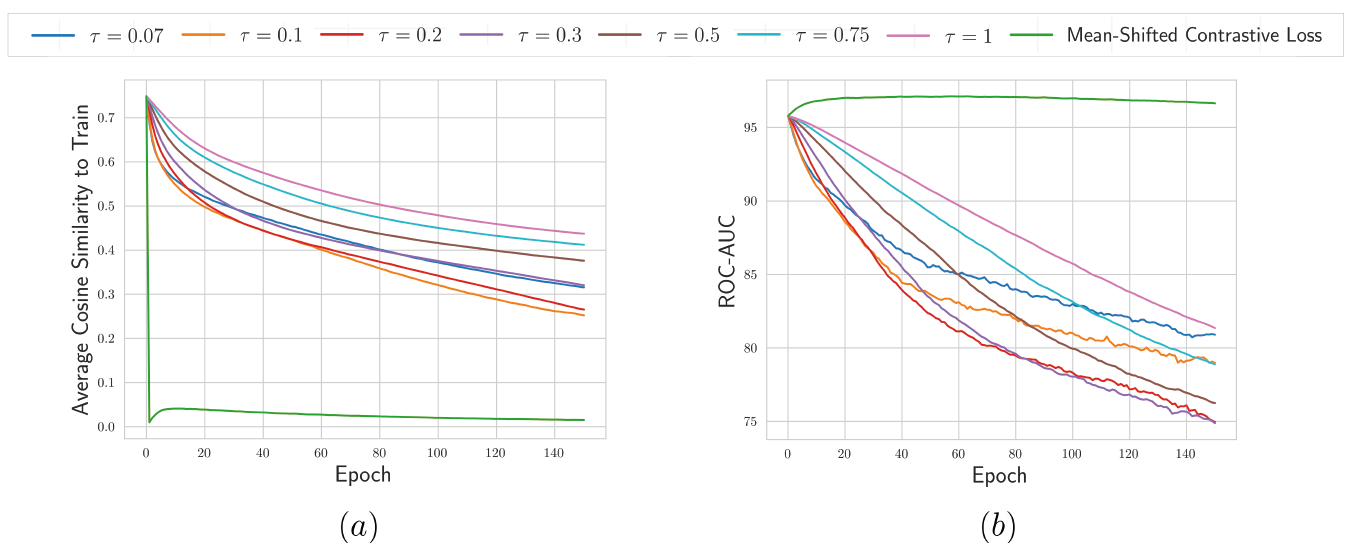}
    \end{tabular}
    \caption{CIFAR-10 "Airplane" class. Ablation study of different temperature parameters while optimizing standard contrastive loss and mean-shifted contrastive loss with $\tau=0.25$. \textbf{\textit{(a)}} Similarity between pairs of images. \textbf{\textit{(b)}} The standard contrastive objective is unsuitable for OCC feature adaptation as it results in very fast catastrophic collapse independently of the chosen $\tau$.}
    \label{appendix:temp_fig}

\end{figure*}

The temperature $\tau$ has an important role in the contrastive objective. It was previously shown by \cite{wang2021understanding} that it influences both the uniformity of sample distribution on the hypersphere and the weight given to hard negative samples. When the temperature approaches infinity, the model pays equal attention to the negative samples and when it approaches zero, the model ignores all the negative samples but the one with the maximum similarity. Based on this analysis, as the temperature increases, the feature space distribution tends to be less uniform, and when $\tau$ is small, the feature space distribution is closer to a uniform distribution. This suggests that using a small temperature parameter while optimizing the standard contrastive objective would solve the optimization dynamics failure that the above suffers from. This in fact not the case, in Fig.~\ref{appendix:temp_fig}.a we present an ablation study of different temperature parameters while optimizing the standard contrastive loss. We observe that using a smaller $\tau$ slightly helps uniformity but not enough to make the optimization focus on improving the features so that they are invariant to augmentations, as catastrophic collapse still occurs (Fig.~\ref{appendix:temp_fig}.b).


\begin{table*}[ht]
  \centering
  \begin{tabular}{lllllllll}
    \toprule
	&	DeepSVDD 	&	MRot 	&	DROC 	&	CSI 	&	PANDA 	&	Ours	\\

    \cmidrule(r){1-1}
    \cmidrule(r){2-7}

0	&	61.7	&	77.5	&	90.9	&	89.9	&	\textbf{97.4}	&	97.0	\\
1	&	65.9	&	96.9	&	98.9	&	\textbf{99.1}	&	98.4	&	98.7	\\
2	&	50.8	&	87.3	&	88.1	&	93.1	&	93.9	&	\textbf{94.8}	\\
3	&	59.1	&	80.9	&	83.1	&	86.4	&	90.6	&	\textbf{94.3}	\\
4	&	60.9	&	92.7	&	89.9	&	93.9	&	\textbf{97.5}	&	96.9	\\
5	&	65.7	&	90.2	&	90.3	&	93.2	&	94.4	&	\textbf{97.2}	\\
6	&	67.7	&	90.9	&	93.5	&	95.1	&	97.5	&	\textbf{98.2}	\\
7	&	67.3	&	96.5	&	98.2	&	\textbf{98.7}	&	97.5	&	98.3	\\
8	&	75.9	&	95.2	&	96.5	&	97.9	&	97.6	&	\textbf{98.5}	\\
9	&	73.1	&	93.3	&	95.2	&	95.5	&	97.4	&	\textbf{98.3}	\\
						
    \cmidrule(r){1-1}
    \cmidrule(r){2-7}

Mean	&	64.8	&	90.1	&	92.5	&	94.3	&	96.2	&	\textbf{97.2}	\\	

    \bottomrule
  \end{tabular}
\caption{CIFAR-10 anomaly detection performance (mean ROC-AUC\%). Bold denotes the best results.}
\label{tab:cifar-10}
\end{table*}

\begin{table*}[ht]
  \centering
  \begin{tabular}{lllllllll}
    \toprule
	&	DeepSVDD &	MRot 	&	DROC &	CSI 	&	PANDA	&	Ours	\\

    \cmidrule(r){1-1}
    \cmidrule(r){2-7}

0	&	66.0	&	77.6	&	82.9	&	86.3	&	91.5	&	\textbf{96.0}	\\
1	&	60.1	&	72.8	&	84.3	&	84.8	&	92.6	&	\textbf{95.3}	\\
2	&	59.2	&	71.9	&	88.6	&	88.9	&	98.3	&	\textbf{98.1}	\\
3	&	58.7	&	81.0	&	86.4	&	85.7	&	96.6	&	\textbf{97.9}	\\
4	&	60.9	&	81.1	&	92.6	&	93.7	&	96.3	&	\textbf{97.6}	\\
5	&	54.2	&	66.7	&	84.5	&	81.9	&	94.1	&	\textbf{96.8}	\\
6	&	63.7	&	87.9	&	73.4	&	91.8	&	96.4	&	\textbf{98.5}	\\
7	&	66.1	&	69.4	&	84.2	&	83.9	&	91.2	&	\textbf{93.4}	\\
8	&	74.8	&	86.8	&	87.7	&	91.6	&	94.7	&	\textbf{97.2}	\\
9	&	78.3	&	91.7	&	94.1	&	95.0	&	94.0	&	\textbf{96.2}	\\
10	&	80.4	&	87.3	&	85.2	&	94.0	&	96.4	&	\textbf{97.1}	\\
11	&	68.3	&	85.4	&	87.8	&	90.1	&	92.6	&	\textbf{96.4}	\\
12	&	75.6	&	85.1	&	82.0	&	90.3	&	93.1	&	\textbf{95.8}	\\
13	&	61.0	&	60.3	&	82.7	&	81.5	&	89.4	&	\textbf{92.6}	\\
14	&	64.3	&	92.7	&	93.4	&	94.4	&	98.0	&	\textbf{99.0}	\\
15	&	66.3	&	70.4	&	75.8	&	85.6	&	89.7	&	\textbf{92.5}	\\
16	&	72.0	&	78.3	&	80.3	&	83.0	&	92.1	&	\textbf{95.2}	\\
17	&	75.9	&	93.5	&	97.5	&	97.5	&	97.7	&	\textbf{98.4}	\\
18	&	67.4	&	89.6	&	94.4	&	95.9	&	94.7	&	\textbf{97.6}	\\
19	&	65.8	&	88.1	&	92.4	&	95.2	&	92.7	&	\textbf{97.0}	\\
						
    \cmidrule(r){1-1}
    \cmidrule(r){2-7}

Mean	&	67.0	&	80.1	&	86.5	&	89.6	&	94.1	&	\textbf{96.4}	\\	

    \bottomrule
  \end{tabular}
\caption{CIFAR-100 coarse-grained version anomaly detection performance (mean ROC-AUC\%). Bold denotes the best results.}
\label{tab:cifar-100}
\end{table*}

\begin{table*}[ht]
  \centering
  \begin{tabular}{lllllllll}
    \toprule
	&	DeepSVDD 	&	MRot	&	DROC 	&	CSI 	&	PANDA 	&	Ours	\\

    \cmidrule(r){1-1}
    \cmidrule(r){2-7}

Cat	&	49.2	&	87.7	&	91.7	&	85.7	&	99.2	&	\textbf{99.4}	\\
Dog	&	51.8	&	84.2	&	87.5	&	86.9	&	95.4	&	\textbf{99.2}	\\
						
    \cmidrule(r){1-1}
    \cmidrule(r){2-7}

Mean	&	50.5	&	86.0	&	89.6	&	86.3	&	97.3	&	\textbf{99.3}	\\	

    \bottomrule
  \end{tabular}
\caption{CatsVsDogs anomaly detection performance (mean ROC-AUC\%). Bold denotes the best results.}
\label{tab:catsdogs}
\end{table*}

\end{document}